\documentclass[preprint,12pt,authoryear]{elsarticle}

\usepackage{amssymb}

\begin{document}

\begin{frontmatter}



\title{Application of deep learning techniques in non-contrast computed tomography pulmonary angiogram for pulmonary embolism diagnosis}



\author{I-Hsien Ting}
\ead{iting@nuk.edu.tw}
\affiliation{organization={National University of Kaohsiung},
city={Kaohsiung},
,country={Taiwan}, 
,email={iting@nuk.edu.tw}
,orcid={0000-0002-6587-2438}
  }

\author{Yi-Jun Tseng}
\affiliation{organization={National University of Kaohsiung},
city={Kaohsiung},
,country={Taiwan},
,email={m1093310@mail.nuk.edu.tw}
  }

\author{Yu-Sheng Lin}
\affiliation{{Chiayi Chang Gung Memorial Hospital},
city={Chiayi},
,country={Taiwan},
,email={dissert@adm.cgmh.org.tw},
,orcid={0009-0003-0862-0511}
}

\begin{abstract}
Pulmonary embolism is a life-threatening disease, early detection and treatment can significantly reduce mortality. In recent years, many studies have been using deep learning in the diagnosis of pulmonary embolism with contrast medium computed tomography pulmonary angiography, but the contrast medium is likely to cause acute kidney injury in patients with pulmonary embolism and chronic kidney disease, and the contrast medium takes time to work, patients with acute pulmonary embolism may miss the golden treatment time.

This study aims to use deep learning techniques to automatically classify pulmonary embolism in CT images without contrast medium by using a 3D convolutional neural network model. The deep learning model used in this study had a significant impact on the pulmonary embolism classification of computed tomography images without contrast with 85\% accuracy and 0.84 AUC, which confirms the feasibility of the model in the diagnosis of pulmonary embolism.

\end{abstract}


\end{frontmatter}


\section{Introduction}
\subsection{Research Background}
The modern lifestyle of individuals often involves prolonged periods of sitting, whether it's for work, commuting, or leisure activities. This sedentary lifestyle is prevalent in students due to educational demands, as well as in the elderly, where reduced mobility can lead to extended periods of inactivity or confinement to bed. These factors, among others such as pregnancy, uncontrolled cardiovascular diseases, and a history of cancer, can contribute to the development of lower limb venous thrombosis. Notably, studies like that of \cite{b01} have shown that lower limb venous thrombosis is the most common cause of Pulmonary Embolism (PE) resulting from blood clots.

In recent years, the field of deep learning has seen significant growth, particularly in applications within the domain of medical imaging. Research focused on employing deep learning to aid in diagnosing PE through computed tomography pulmonary angiography (CTPA) has gained prominence. Studies like that of \cite{b02} have sought to replace manual interpretation of CTPA results with deep learning techniques, aiming to alleviate the clinical workload for medical practitioners. Additionally, \cite{b03} have improved the quality of medical images from CTPA scans, thus reducing the radiation dose.

When a physician strongly suspects a patient may have PE, CTPA has become the primary diagnostic method (\cite{b04}). It enables immediate identification and prompt treatment of patients based on medical imaging results. Nevertheless, the risk of underdiagnosing PE remains a significant concern (\cite{b05}), mainly because of the need for physicians to manually interpret CTPA images. Each patient's CTPA scan typically comprises approximately 100-300 slices, necessitating extensive time for physicians to meticulously examine each pulmonary artery to detect potential embolic symptoms. Lack of experience, physician fatigue (\cite{b06}), or inadequate physician coverage can lead to prolonged judgment times and diagnostic errors. Consequently, the healthcare system is burdened with increased stress (\cite{b07}, \cite{b08}). Moreover, as the application of CTPA scans grows, ensuring the provision of accurate and timely diagnostic images becomes increasingly challenging for healthcare systems and physicians (\cite{b09}).

Patients undergoing CTPA scans are at risk of developing acute kidney injury (AKI), a common occurrence in critically ill patients with PE (\cite{b10}). Once AKI occurs, the patient's risk of death rises significantly. Additionally, CTPA scans require the administration of contrast medium, and this requires a certain period to allow the contrast medium to circulate through the pulmonary system before the actual CT scan (\cite{b11}). Patients with acute PE may lose valuable treatment time while waiting for the contrast medium to take effect. This delay could be critical since patients with acute PE require immediate diagnosis and treatment.

\subsection{Research Motivation}
CTPA images not only serve as the most commonly used method for diagnosing PE but also allow physicians to definitively determine the presence of embolic symptoms in patients without the need for further diagnostic confirmation. This makes CTPA images ideal for supervised learning methods in deep learning for image classification (\cite{b12}). In recent research, deep learning algorithms used in the medical imaging field predominantly employ supervised learning (\cite{b13}), particularly convolutional neural networks (CNNs) (\cite{b14}). To address healthcare system and physician-related issues, Yang et al.'s research (\cite{b15}) proposed a two-stage CNN for assessing PE in CTPA scans. The model exhibited a sensitivity of 75.4\% when evaluated using CTPA test data from 20 patients. \cite{b02} introduced PENet, a 3D CNN, for detecting PE. When tested on two datasets, PENet achieved AUROCs of 0.84 and 0.85. These methods represent new avenues for human-robot collaboration (HRC) in healthcare (\cite{b16}). They reduce the probability of diagnostic errors while also decreasing the time physicians spend interpreting images, enabling more appropriate treatment for patients.

\begin{figure*}
    \centering
    \includegraphics[width=0.8\textwidth]{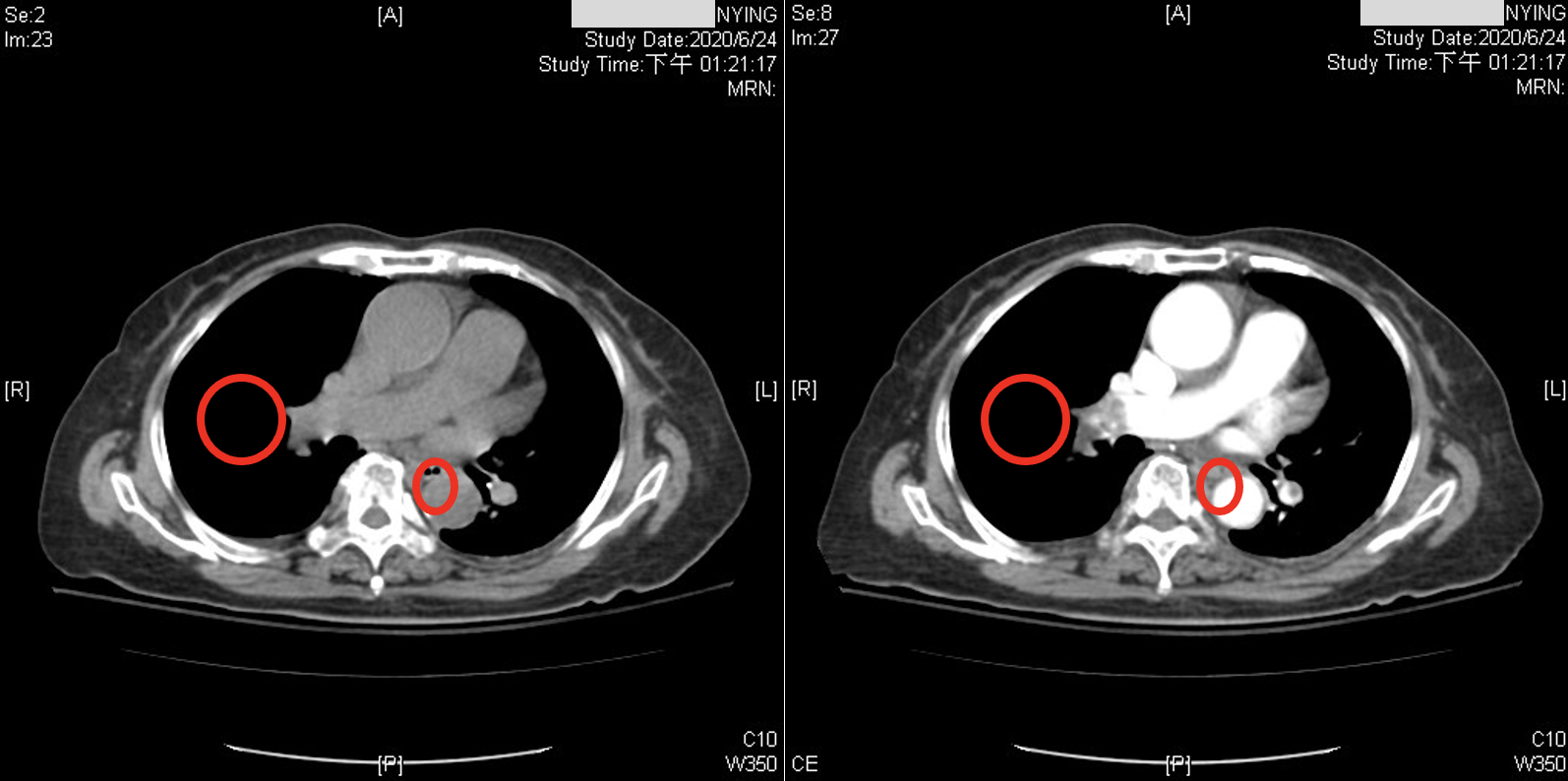}
    \caption{Illustrates the difference between CTPA scans with and without contrast medium: the left image is without a contrast medium, while the right image is with a contrast medium. In the right image, the circled area clearly shows a darker region. This is because PE causes blockages in the pulmonary vessels, resulting in reduced blood circulation and lower oxygen levels within these vessels. The contrast medium enhances the visibility of the occlusion (darker area). In contrast, the left image, without a contrast medium, makes it more challenging to visually determine the presence of a PE.}
    \label{fig:img1}
\end{figure*}

While recent studies have used CNNs to diagnose PE in CTPA. Although the accuracy could be improved from a physician's point of view, these images have been acquired after the administration of contrast medium. CTPA images with contrast medium tend to provide clearer distinctions for diagnosis, as shown in Figure 1. However, contrast medium can have adverse effects on the body. They exert cytotoxic effects on the renal tubular epithelial cells (RTE) of the kidneys, leading to loss of function, severe cell damage, and even cell death (\cite{b17}). Furthermore, factors such as patient history and prior surgeries can influence the impact of contrast medium on the kidneys (\cite{b18}). Studies have confirmed that contrast medium are the most significant independent risk factor for acute kidney injury in patients with severe chronic kidney disease (\cite{b19}). Additionally, the quality of images can be influenced by factors such as the time the contrast medium circulates in the patient's body, the patient's physiological factors, and the concentration of the contrast medium (\cite{b11}). In summary, the risk associated with CTPA images is higher than non-contrast CT scans. This is because CTPA scans involving contrast medium introduce the risk of renal exposure to the contrast medium, increasing the overall risk for patients requiring related medical procedures. Consequently, this study aims to apply deep learning to classify PE in non-contrast CT scans of the lungs.

\subsection{Research Objective}
Taking into consideration the aspects discussed above, this study will use data obtained from our collaborating institution to perform deep learning classification. The study aims to automatically classify the presence or absence of PE in non-contrast CT scans of the lungs using a deep learning CNN model. Through the research objectives outlined above, we hope to employ the results of this study to utilize supervised learning techniques in deep learning, enhancing the automated detection of PE. By doing so, we aim to alleviate the burden on medical practitioners, improve the efficiency of PE diagnosis, reduce diagnostic errors, and minimize the physical burden on patients, ultimately lowering mortality rates.

\section{Literature Review}
\subsection{Pulmonary Embolism}
Pulmonary Embolism (PE) is a condition characterized by the blockage of any pulmonary blood vessel by a blood clot. This obstruction hinders the flow of blood within the vessel, resulting in reduced oxygen content in the lung tissue, leading to hypoxemia, and posing a life-threatening risk (\cite{b20}). PE is a common presentation in emergency departments and a prevalent cause of mortality. Annually, approximately 50,000 to 200,000 people succumb to PE, with a 30-day mortality rate ranging from 0.5\% to over 20\%, contingent upon the presentation at the time of onset (\cite{b21}). Typical symptoms of PE include dyspnea, chest pain, cough, and other common manifestations. However, less common symptoms such as gradual-onset dyspnea over days to weeks or fainting with less prominent respiratory symptoms can also occur. Therefore, healthcare professionals need to conduct further examinations on patients with potential cardiopulmonary symptoms to confirm the presence of PE. Delayed diagnosis of PE can have severe consequences, with up to 95\% of patients succumbing before diagnosis, and most fatalities occurring in untreated individuals (\cite{b22}, \cite{b23}). Therefore, early detection and treatment of PE can significantly reduce mortality rates.
\subsection{Contrast Medium}
Contrast medium is a pharmaceutical agent used in medical imaging to enhance the visibility of body tissues. computed tomography pulmonary angiography (CTPA) is a medical imaging technique that employs contrast medium. CTPA is the primary method for diagnosing PE (\cite{b04}). It allows physicians to promptly identify and treat PE using CTPA images. In recent years, deep learning techniques have been applied to diagnose PE using CTPA images, such as the studies conducted by \cite{b02} , \cite{b24}, and \cite{b15}. Although CTPA is the primary diagnostic method for PE, the use of contrast medium carries potential risks for the patient. Research by \cite{b17} indicates that contrast medium exerts cytotoxic effects on renal tubular epithelial cells (RTE) in the kidneys, leading to functional impairment, severe cellular damage, and cell apoptosis and necrosis. \cite{b18} proposed that factors related to the patient and surgical history significantly impact the risk of developing Acute Kidney Injury (AKI) following contrast medium administration. In patients with chronic kidney disease, contrast medium is the most potent independent risk factor for acute kidney injury \cite{b19}. Given the elevated risk of contrast medium to the kidneys, it is prudent to explore the application of deep learning techniques to non-contrast CT scans of the lungs for the diagnosis of PE. Currently, there is a lack of relevant studies in this area. Thus, this study will focus on utilizing non-contrast CT scans of the lungs in conjunction with deep learning techniques to classify the presence or absence of PE.
\begin{figure*}
    \centering
    \includegraphics[width=1\textwidth]{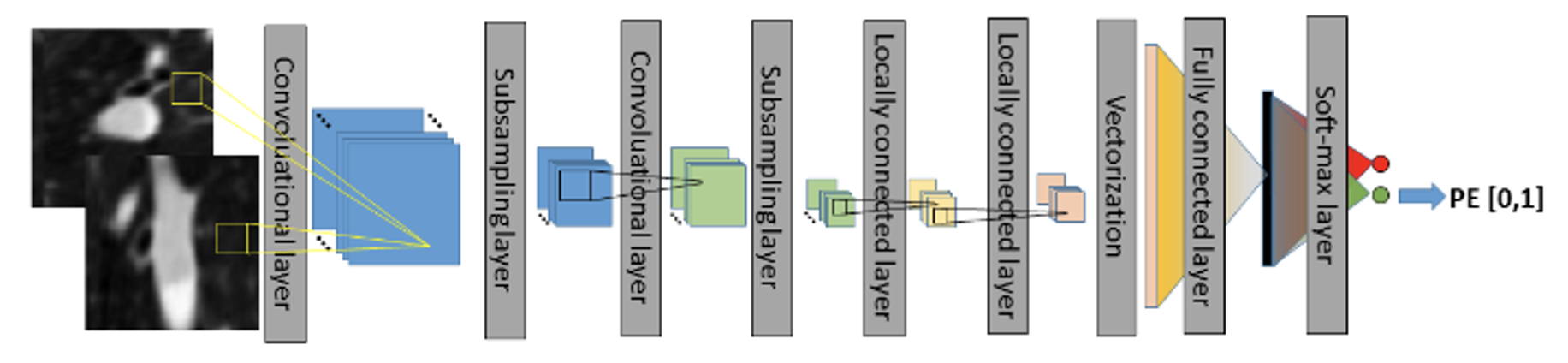}
    \caption{CNN model architecture by \cite{b24}}
    \label{fig:img2}
\end{figure*}
\subsection{Application of Deep Learning}
In recent years, the majority of algorithms in medical imaging research have employed convolutional neural networks (CNNs), primarily in the diagnosis of PE using CTPA images. For example, \cite{b24} applied a 3D CNN to CTPA images for the detection of PE. They initiated the process with lung segmentation, followed by the identification of potential embolic regions within the lungs using a tobogganing algorithm. They then represented these regions using multi-planar vessel-aligned images and generated 2-channel images for each potential embolic region. Finally, a 9-layer CNN model, as shown in Figure 2, was employed for classification, including two convolutional layers, subsampling layers, two locally connected layers, vectorization, and fully connected layers with a Softmax layer. The convolutional layers and subsampling layers focused on feature extraction, followed by using locally connected layers to extract regionally relevant features to enhance computational efficiency. The 2D images were then vectorized for classification using fully connected and Softmax layers. This approach involves substantial pre-processing to generate features as inputs for the CNN model.

\begin{figure*}
    \centering
    \includegraphics[width=0.9\textwidth]{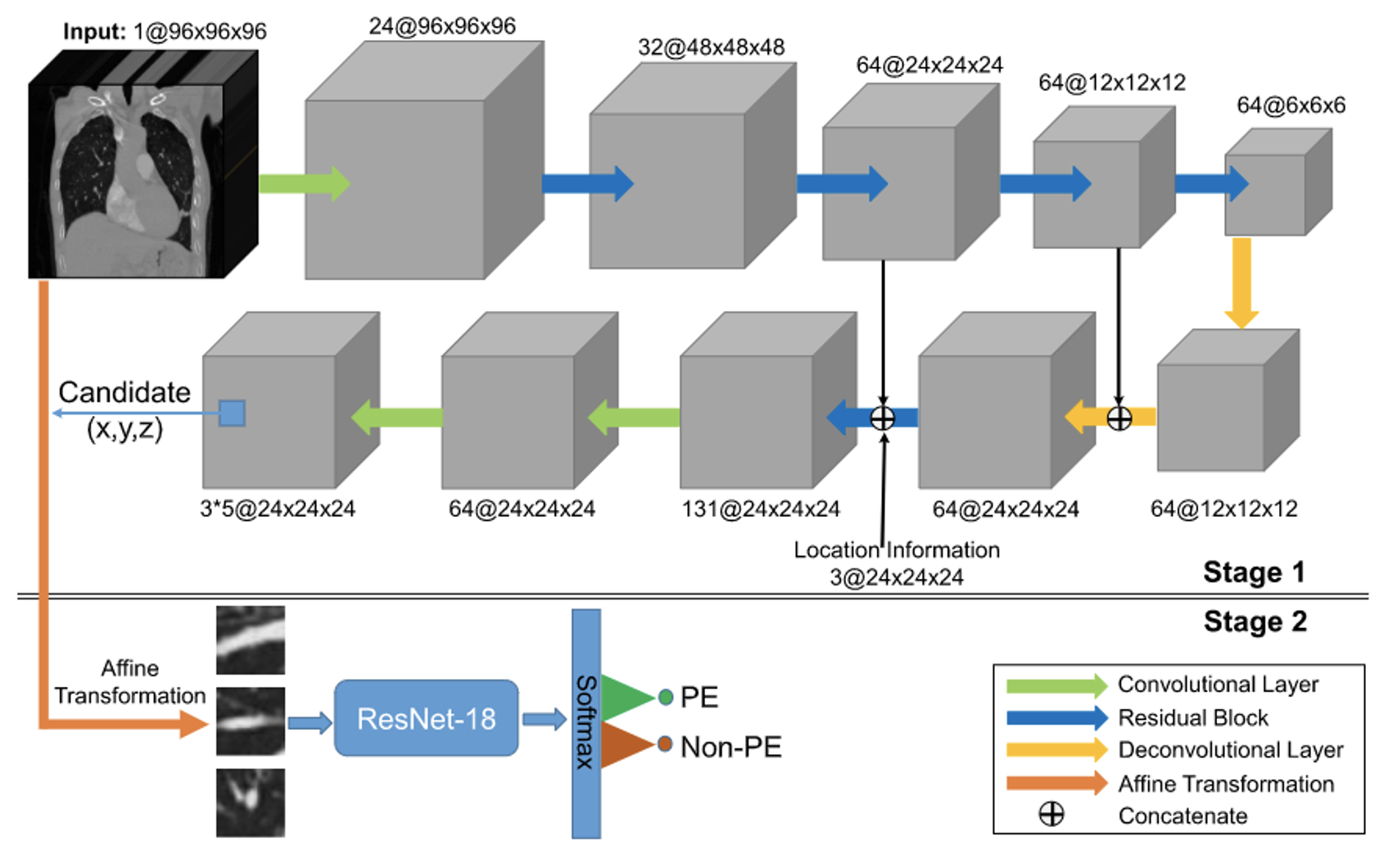}
    \caption{CNN model architecture by \cite{b15}}
    \label{fig:img3}
\end{figure*}

\cite{b15} proposed a two-stage CNN model, evaluated using a test set of CTPA images from 20 patients. They achieved a sensitivity of 75.4\%. In the first stage, they used a 3D fully convolutional network (FCN) to segment and extract potential pulmonary embolic regions, as illustrated in Figure 3. The encoder began with a 3D convolutional layer, followed by a max-pooling layer and four residual blocks for feature extraction. A residual block and two deconvolution layers were used to up-sample the feature image, incorporating location information regarding thrombus. A skip connection linked the final two residual blocks of the encoder with their counterparts in the decoder. In the second stage, the extracted pulmonary embolic regions underwent vessel alignment using ResNet-18 to assess obstruction in the contrast-enhanced blood vessels, thereby identifying PE symptoms. This approach focused on segmenting critical regions and extracting potential lesions, rather than utilizing the entire CTPA image for evaluation.

\begin{figure*}
    \centering
    \includegraphics[width=0.9\textwidth]{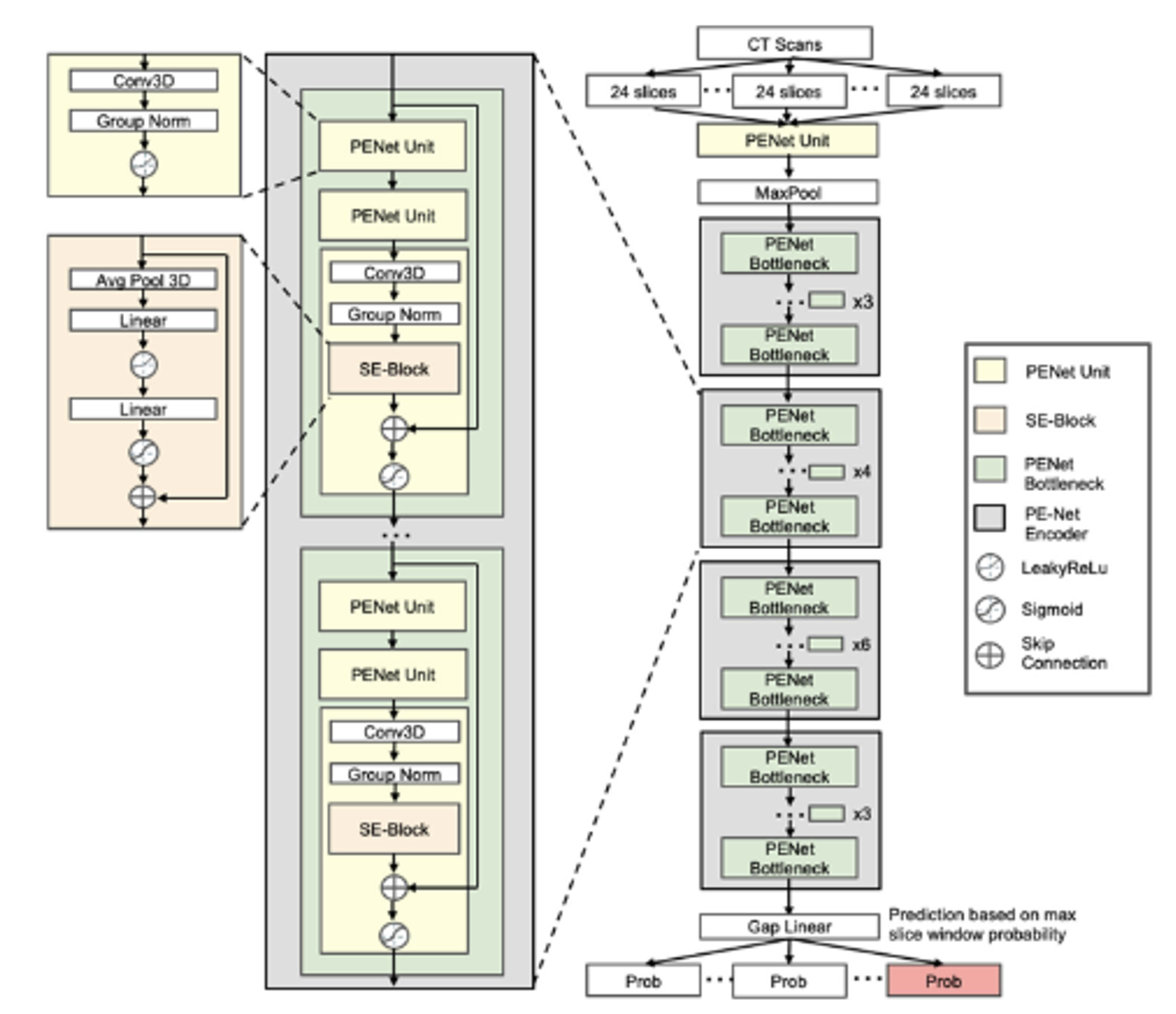}
    \caption{CNN model architecture by \cite{b02}}
    \label{fig:img4}
\end{figure*}

\cite{b02} introduced the PENet, a 3D end-to-end CNN model with a two-test set evaluation for PE detection, achieving AUROC values of 0.84 and 0.85. They developed a PENet composed of four modules, as depicted in Figure 4: the PENet unit, Squeeze-and-Excitation block (SE-block), PENet bottleneck, and the PENet encoder. The PENet unit processed 3D input data through 3D CNN and group normalization, utilizing LeakyReLU activation. The SE-block, by simulating interdependencies among channels, autonomously adjusted the feature characteristics of the channels. The PENet bottleneck consisted of three PENet units, with the final unit inserting an SE-block after group normalization. A residual connection was applied between the input of the PENet bottleneck and the output of the SE-block. The model was concluded with a PENet unit, four PENet encoders, and Gap Linear activation.

\begin{figure*}
    \centering
    \includegraphics[width=0.6\textwidth]{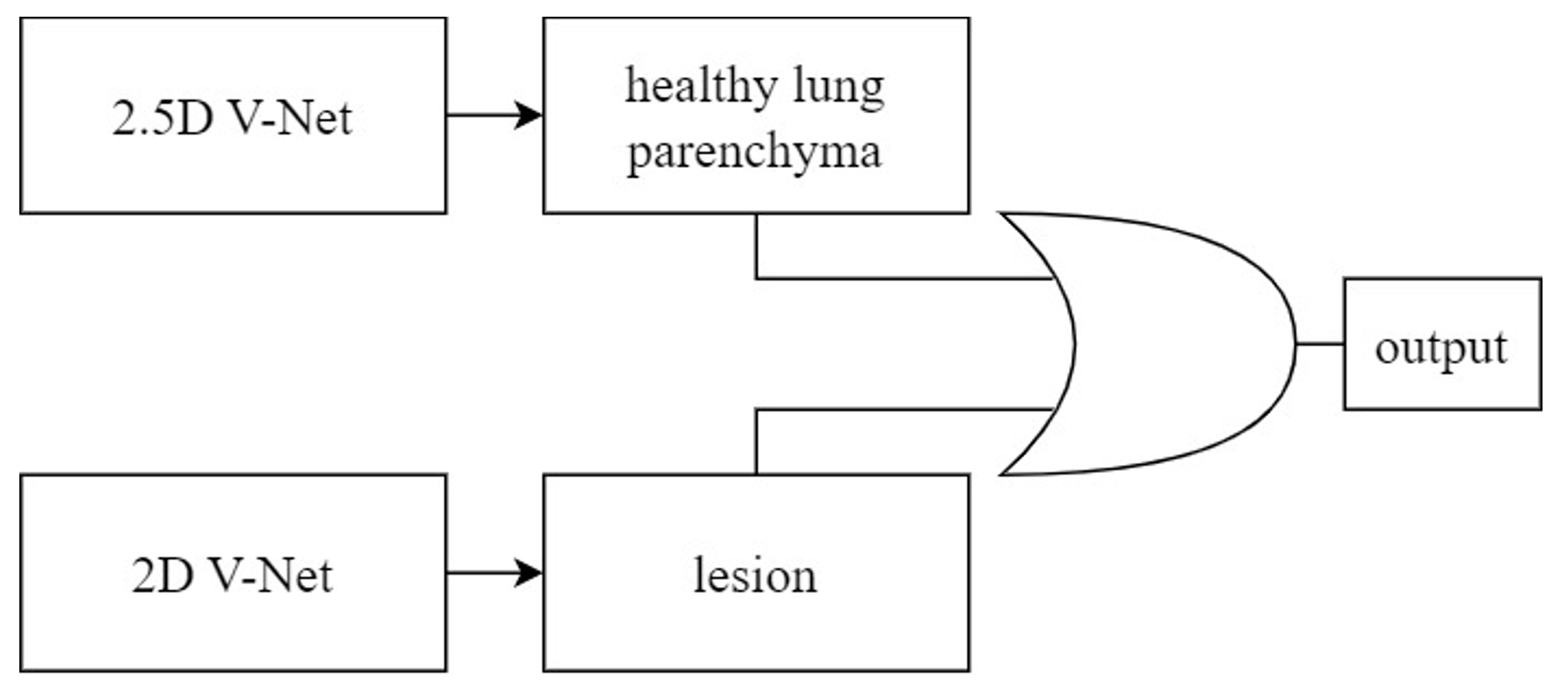}
    \caption{Workflow for lung parenchyma and lesion segmentation by \cite{b25}}
    \label{fig:img5}
\end{figure*}

These studies focused on applying CNN models to CTPA images. Some have mentioned the desire to utilize non-contrast CT scan images. For instance, \cite{b25} employed non-contrast CT lung images to identify lung parenchyma and lesion-related manifestations in COVID-19 patients. This research involved segmenting regions of interest (ROI) and extracting features from these regions within the complete CT scan images of each patient. To correctly segment the lung regions of COVID-19 patients, they trained two CNN models based on the V-Net architecture, including a healthy lung parenchyma model and a lesion model, as depicted in Figure 5. These models used OR logic to produce the final outcome. This method focused on segmenting ROI, extracting potential pathological regions, and not using the entire CT scan lung images. This study, however, employed a more intricate approach, involving training two CNN models to correctly segment the lung regions of COVID-19 patients. It employed substantial pre-processing to generate features as inputs for the CNN model and was more complex compared to other studies.

Therefore, this study aims to apply CNN models to non-contrast CT scan images of the lungs. The dataset for lung CT scan images will be provided by collaborating institutions. The reason for not using existing datasets is that they exclusively comprise CTPA images, making them unsuitable for this research. Thus, the study will use the dataset provided by collaborating institutions.

\section{Research Methodology}
\subsection{Research Framework}
The research framework of this study is depicted in Figure 6. The study utilizes a dataset of computed tomography (CT) scans of the lungs provided by collaborating institutions. This dataset is divided into a training set, a validation set, and a test set. Pre-processing is performed on the CT scan images, including adjustment of Hounsfield Units (HU) ranges and image resizing. Subsequently, the pre-processed data is used for model training. The model is then evaluated and analyzed using the validation set. If the results require improvement, model parameters are adjusted. Finally, the test set is used for classifying the presence or absence of Pulmonary Embolism (PE), and the results are reported.

\begin{figure*}
    \centering
    \includegraphics[width=1\textwidth]{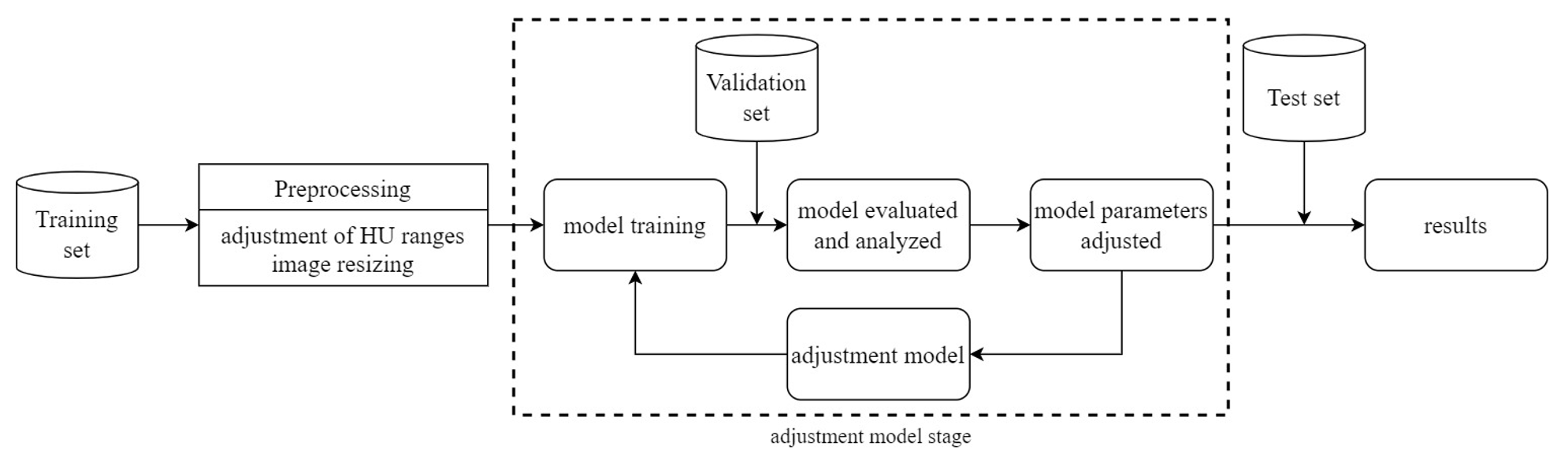}
    \caption{Research framework}
    \label{fig:img6}
\end{figure*}

\subsection{Data Source}

This study utilizes a dataset provided by collaborating institutions. These datasets are sourced from the radiology department of these institutions and have been subjected to review by an Institutional Review Board (IRB) in accordance with the Personal Data Protection Act. CT scans involve the combination of X-rays and computer technology to produce cross-sectional images of the body. These images can be further processed to create detailed 3D images. Each CT scan slice has a thickness of 1 millimeter, and each patient's scan typically consists of 100 to 300 slices, as shown in Figure 7. In total, data has been collected from 192 patients and divided into training (138 patients), validation (34 patients), and test sets (20 patients), as shown in Table 1.

\begin{figure*}
    \centering
    \includegraphics[width=0.7\textwidth]{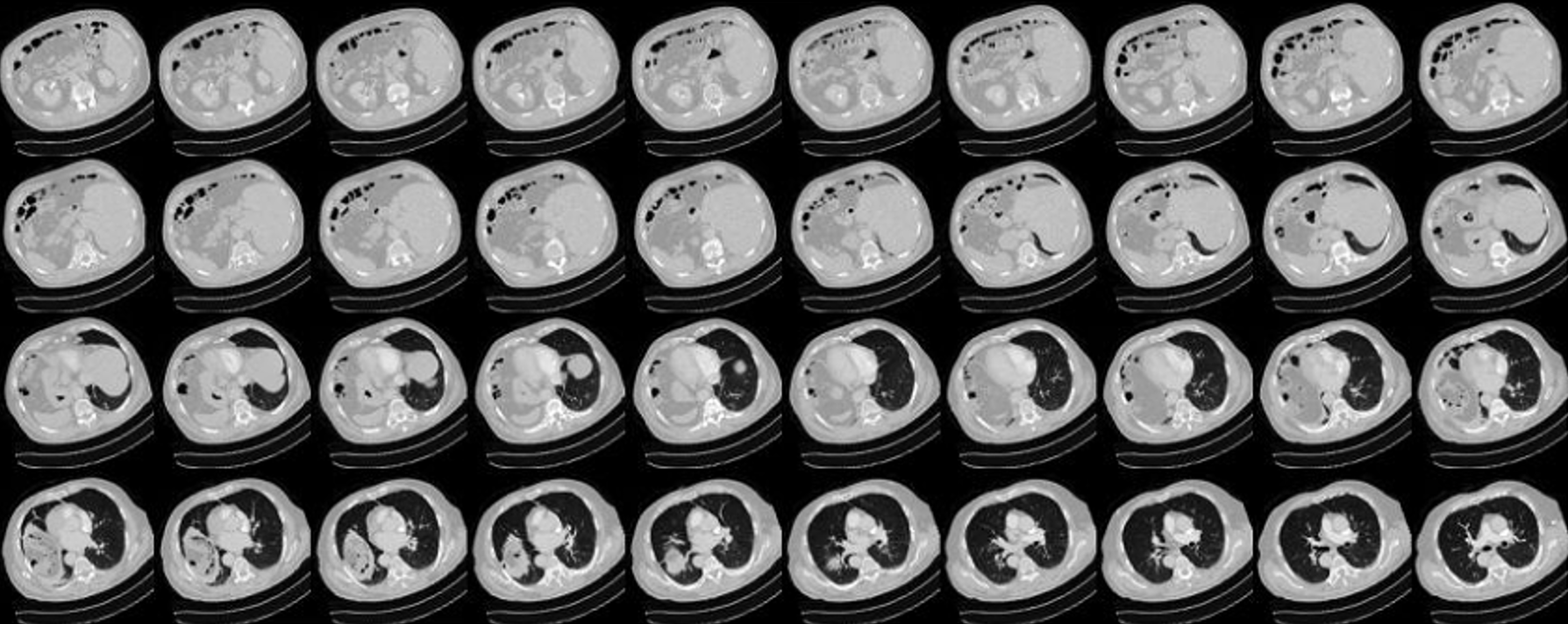}
    \caption{CT slice of the lung. CT slices of a patient's lungs, 40 of which were taken and visualized}
    \label{fig:img7}
\end{figure*}

\begin{table*}
\caption{\label{tbl1}Data set of collaborating institutions}
\centering 
\begin{tabular}{lllll}
\hline
& Training & Validation & Testing & Overall\\
\hline
Number of Studies & 138 & 34 & 20 & 192\\
\hline
Number of positive PE & - & - & - & 80\\
\hline
Number of negative PE & - & - & - & 112 \\
\hline
\end{tabular} 
\end{table*}

\subsection{Data Pre-processing}

\subsubsection{Adjusting Hounsfield Units}
CT scans measure the X-ray absorption capabilities, known as radiodensity, of different body tissues. These scans produce grayscale images where tissue density is represented as shades of gray. The HU scale ranges from -1000 (black on the grayscale, representing air) to +3000 (the upper limit, representing dense bone), as mentioned by reference \cite{b26}. HU are measurement values used by radiologists to interpret CT scan images, aiding in image analysis and disease diagnosis \cite{b27}. Table 2 below displays the HU ranges utilized in this study.

\begin{table*}
\caption{\label{tbl2}HU for Different Substances}
\centering 
\begin{tabular}{ll}
\hline
Substance&HU\\
\hline
Air&-1000\\
\hline
Lung&-700to-600\\
\hline
Non-coagulated blood&+13 to +50\\
\hline
Coagulated blood&+50 to +75\\
\hline
Vascular calcification&$\geq$+130\\
\hline
Cancellous bone&+300 to +400\\
\hline
Compact bone&+3000\\
\hline

\end{tabular} 
\end{table*}

As Table 2 illustrates, the HU range for lung tissue extends from -1000 to +3000. Therefore, this study adjusts the HU of the CT scan images to several specific ranges to explore their impact on model performance:

\begin{enumerate}
    \item 	-1000 to +80: This range encompasses air, lung tissue, non-coagulated blood, and distinguishes coagulated blood at its maximum HU value.
\item -1000 to +130: Adjusting the range to +130 extends the observable coagulated blood range while still excluding vascular wall calcification.
\item -1000 to +270: In this range, additional lung tissue and vascular wall calcification features are observable.
\item -1000 to +400: This relatively broad range covers various lung tissue densities, including bones with different radiodensity, with +400 set as the upper limit.
\item -1000 to +800: The widest HU range in this study covers all possible tissue and material types, including lung tissue and vascular calcification.
\end{enumerate}

In addition to adjusting the range from -1000 HU for air, this study also considered the detection of PE in the lungs, so the range was adjusted from -700 HU for lungs, and the following ranges were adjusted:

\begin{enumerate}
\item  -700 to +80
\item  -700 to +130
\item -700 to +270
\item -700 to +400
\item -700 to +800
\end{enumerate}

The reasons for the above range adjustments are the same as those for the Heinz units starting at -1000, so we will not repeat them.
To aid the model in better understanding and processing HU, this study normalizes the HU values. If HU values exceed the maximum (minimum) value, they are displayed as the input's maximum (minimum) value. These HU range adjustments aim to facilitate model interpretation by scaling HU values to a range between 0 and 1.


\subsubsection{The Architecture of the CNN model}
This study employs a 3D CNN model based on the architecture proposed by \cite{b28}. As depicted in Figure 8, the model comprises four 3D convolutional layers. The first two layers contain 64 filters, followed by layers with 128 and 256 filters. All filters have a kernel size of 3×3×3. Each convolutional layer is succeeded by a max-pooling layer with a stride of 2, ReLU activation, and Batch Normalization. These layers serve various purposes, including feature extraction, feature condensation, increasing learning rates, and preventing overfitting.

\begin{figure*}
    \centering
    \includegraphics[width=0.5\textwidth]{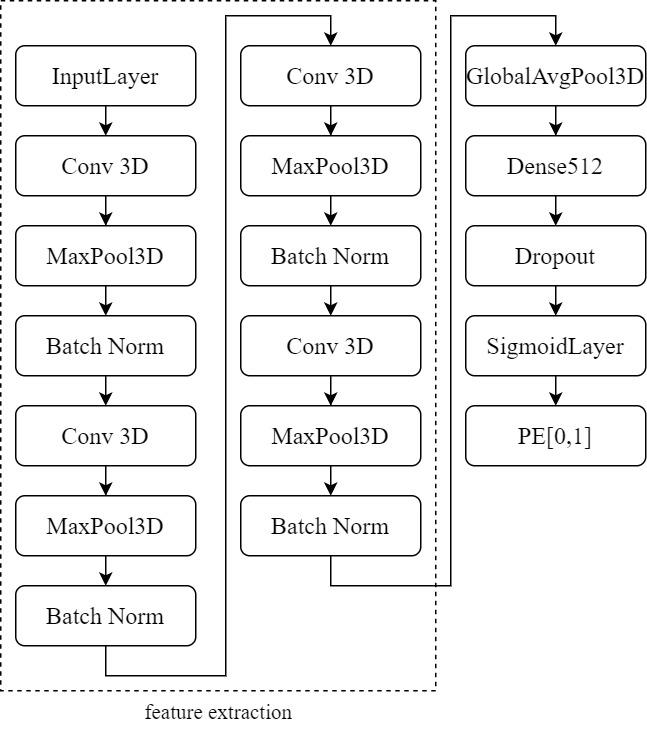}
    \caption{The Architecture of the CNN model}
    \label{fig:img8}
\end{figure*}

The extracted features are processed through Global Average Pooling (GAP), a layer that requires no parameter optimization, thus mitigating the risk of overfitting. After GAP, the input is directed to a Dense layer with 512 neurons. Additionally, the Dropout layer is configured with a dropout probability (p) set to 0.3. This means that each neuron has a 30\% chance of deactivation, which translates to approximately 300 out of 1000 neurons being deactivated. In other words, the output of these 300 neurons is set to 0, effectively addressing the issue of overfitting. Finally, the output results are passed to a Sigmoid layer for binary classification, determining the presence or absence of PE. This model maintains relative simplicity to prevent overparameterization, with only 1,352,897 trainable parameters. Furthermore, due to limitations in training sample size and hardware performance, the model is kept straightforward.

\section{Empirical Study and Evaluation}
\subsection{Experimental Procedure}
The experimental procedure of this study is illustrated in Figure 9.

\begin{figure*}
    \centering
    \includegraphics[width=0.5\textwidth]{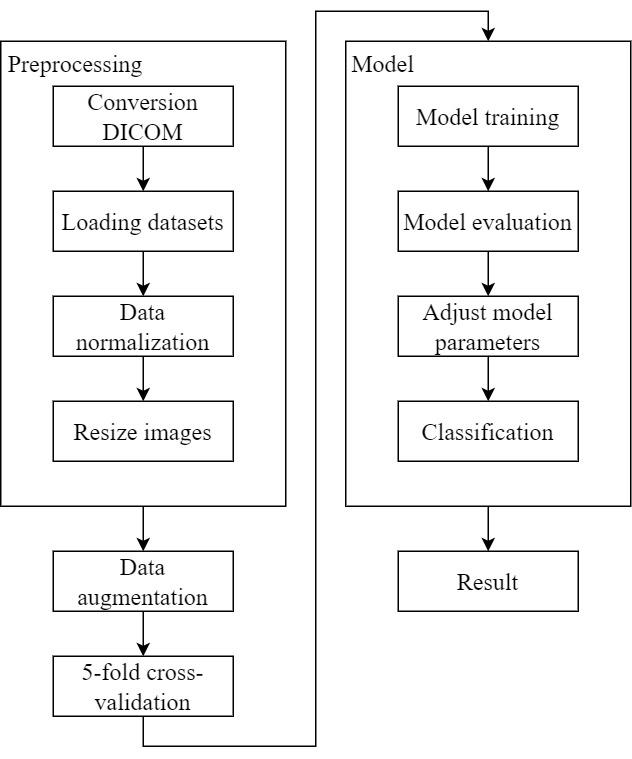}
    \caption{Flowchart of the Experimental Procedure}
    \label{fig:img9}
\end{figure*}

\subsubsection{Data pre-processing}

\begin{enumerate}
\item Conversion from DICOM to NIFTI: A patient's lung computed tomography (CT) scan comprises many DICOM files, which are individual 2D slices. However, NIFTI files combine these 2D images into a 3D volume, making them more convenient for use in deep learning compared to numerous DICOM files.
\item Data normalization: This study adjusted the Hounsfield Units (HU) of lung CT scan images to various ranges, including -1000 to +80, -1000 to +130, -1000 to +270, -1000 to +400, -1000 to +800, and -700 to +80, -700 to +130, -700 to +270, -700 to +400, -700 to +800, and scaled the HU between 0 and 1.
\item Resize images: As described in Chapter 3, Section 3, this study resized all lung computed tomography (CT) scan images to a uniform size for subsequent model training.
\end{enumerate}

\subsubsection{Data augmentation}
Perform data augmentation: Data augmentation was performed at six different angles: -20 degrees, -10 degrees, -5 degrees, 5 degrees, 10 degrees, and 20 degrees, resulting in a sixfold increase in data volume to enhance both sample size and diversity.
\subsubsection{Data splitting}
5-fold cross-validation: As described in Chapter 4, Section 2, the dataset was divided into training and validation sets using 5-fold cross-validation to ensure model training and evaluation accuracy.
\subsubsection{Model training}
\begin{enumerate}
\item Utilizing 3D convolutional neural network (CNN) model: As described in Chapter 3, this study employed a model architecture based on Zunair et al. for model training.
\item Model training: Training was conducted using the training sets obtained through 5-fold cross-validation.
\end{enumerate}

\subsubsection{Model evaluation}
\begin{enumerate}
\item Evaluate model performance: The training and validation sets from 5-fold cross-validation were used to evaluate the model based on accuracy and loss functions.
\item Adjust model parameters: If the model evaluation results require improvement, adjustments to model parameters were made to enhance performance.
\end{enumerate}

\subsubsection{Classification and result reporting}
\begin{enumerate}
\item Test set classification: The test set was used to classify the presence or absence of Pulmonary Embolism (PE).
\item Generate result reports: Classification results were compiled, and the model's overall performance on the test set was evaluated using four metrics from the confusion matrix: accuracy, sensitivity, specificity, and Area Under the Curve (AUC). Experimental results were presented.
\end{enumerate}

\subsection{Evaluation Metrics}

\begin{table*}
\caption{\label{tbl3}The Confusion matrix}
\centering 
\begin{tabular}{lll}
\hline
&Actually Positive&Actually Negative\\
\hline
Predicted Positive&True Positive&False Positive\\
\hline
Predicted Negative&False Negative&True Negative\\
\hline
\end{tabular} 
\end{table*}

This study employs a 5-fold cross-validation approach for model validation and evaluation. Additionally, this study evaluates the model using a confusion matrix. A confusion matrix is a common method for assessing model performance. As shown in Table 3, "True" and "False" represent whether the prediction itself is correct or incorrect, while "Positive" and "Negative" indicate the direction of the prediction, whether it is positive or negative. The confusion matrix consists of four elements: True Positive (TP), True Negative (TN), False Positive (FP), and False Negative (FN). TP refers to correctly predicting positive samples, meaning that the prediction matches the actual situation. TN refers to correctly predicting negative samples, where the prediction matches the actual situation. FP occurs when a negative sample is incorrectly predicted as positive, indicating a mismatch between the prediction and the actual situation. FN refers to incorrectly predicting a positive sample as negative, again indicating a mismatch between the prediction and the actual situation.

Using the elements TP, TN, FP, and FN from the confusion matrix, four key metrics, including accuracy, sensitivity, specificity, and AUC (Area under the Curve), are calculated to evaluate the model. Accuracy represents the ratio of correct predictions. In the medical field, two important metrics for determining whether a patient has a condition are sensitivity and specificity. Sensitivity measures the ability to detect TP samples, i.e., how many individuals with the condition are correctly identified. Specificity measures the ability to detect TN samples, i.e., how many healthy individuals are correctly identified as such. AUC is the area under the ROC curve (Receiver Operating Characteristic Curve), as depicted in Figure 10. The ROC curve's horizontal axis represents 1 Specificity (FPRate), while the vertical axis represents Sensitivity (TPRate). TPRate indicates the proportion of correctly identified positive samples among all actual positive samples. FPRate indicates the proportion of falsely identified positive samples among all actual negative samples. Achi eving a higher TPRate than FPRate signifies better model performance, and AUC helps mitigate issues related to imbalanced sample sizes. The Three indicators are defined as follows.

\begin{equation}
    Accuracy=\frac{TP+TN}{TP+TN+FP+FN} 
\end{equation}

\begin{equation}
    Sensitivity=\frac{TP}{TP+FN} 
\end{equation}

\begin{equation}
    Specificity=\frac{TN}{TN+FP} 
\end{equation}

\begin{figure*}
    \centering
    \includegraphics[width=0.5\textwidth]{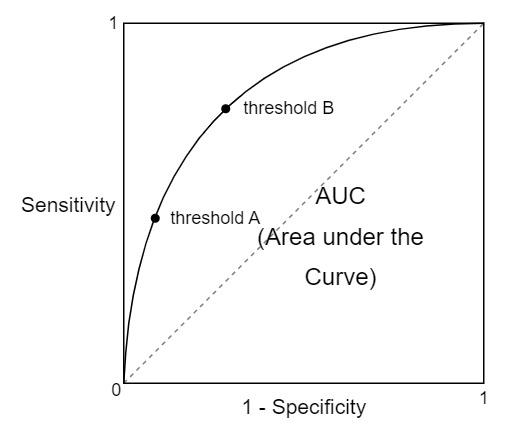}
    \caption{The ROC curve}
    \label{fig:img10}
\end{figure*}

\subsection{Experimental Results}
This section presents the experimental results and evaluates the model's performance across different HU ranges. The study uses 5-fold cross-validation during the experimentation to ensure result accuracy.

The experimental results are shown in Table 4 and 5. In summary, each HU range has its advantages and disadvantages, and a comprehensive comparison of these results can provide a better understanding of the model's strengths and limitations within different HU ranges. The study can then select the most suitable HU range to achieve optimal model performance, considering practical application scenarios, medical requirements, and dataset characteristics.

\begin{table*}
\caption{\label{tbl4}Results for d ifferent HU r anges S tarting from -1000 HU}
\centering 
\begin{tabular}{lllll}
\hline
HU&Accuracy&Sensitivity&Specificity&AUC\\
\hline
-1000 to +80&77\%&67\%&71\%&0.84\\
\hline
-1000 to +130&85\%&33\%&94\%&0.84\\
\hline
-1000 to +270&85\%&33\%&94\%&0.84\\
\hline
-1000 to +400&80\%&33\%&82\%&0.80\\
\hline
-1000 to +800&78\%&67\%&71\%&0.77\\
\hline
\end{tabular} 
\end{table*}

\begin{table*}
\caption{\label{tbl5}Results for different HU ranges: Starting from -700 HU}
\centering 
\begin{tabular}{lllll}
\hline
HU&Accuracy&Sensitivity&Specificity&AUC\\
\hline
-700 to +80&80\%&67\%&76\%&0.75\\
\hline
-700 to +130&83\%&67\%&82\%&0.80\\
\hline
-700 to +270&75\%&67\%&65\%&0.82\\
\hline
-700 to +400&85\%&33\%&94\%&0.76\\
\hline
-700 to +800&73\%&67\%&59\%&0.82\\
\hline
\end{tabular} 
\end{table*}

\section{Conclusion}
\subsection{Conclusion}
Modern lifestyles, characterized by sedentary behaviors, have led to an increased prevalence of sedentary lifestyle-related health risks. Among these risks is the potential for Pulmonary Embolism (PE). In recent years, deep learning has seen rapid advancements in various fields, with the medical sector witnessing research employing deep learning for assisting in the diagnosis of PE. Traditionally, the manual identification of PE by physicians could be time-consuming and prone to errors due to fatigue. As a result, deep learning techniques have been employed to automatically identify the presence of PE in chest computerized tomography (CT) scans. However, most existing studies have used deep learning models for classifying PE in CT pulmonary angiogram (CTPA) images that involve the use of a contrast medium. This study, on the other hand, employs non-contrast CT scans and deep learning models for PE classification.

Differing from previous approaches, this study aims to explore the performance of deep learning models from a different perspective. The absence of a contrast medium in non-contrast chest CT scans often results in less conspicuous PE images compared to contrast images. Therefore, various Hounsfield Unit (HU) range adjustments were made to assess the impact of each range on the performance of a 3D convolutional neural network(CNN) model. The results presented by the model also indicate variations in performance between each HU range. Among these, the HU ranges of -1000 to +130, -1000 to +270, and -700 to +400 exhibited better performance in detecting true negatives, with accuracy rates of 85\%, sensitivity and specificity at 33\% and 94\%, and AUC values of 0.84 for all three. Conversely, the HU ranges of -1000 to +80, -1000 to +800, -700 to +80, and -700 to +130 showed relatively better performance in detecting true positives, with performance in detecting true negatives falling in between. These four ranges demonstrated accuracy rates of 77\%, 78\%, 80\%, and 83\%, sensitivity of 67\%, specificity at 71\%, 71\%, 76\%, and 82\%, and AUC values of 0.84, 0.77, 0.75, and 0.80, respectively.


\subsection{Research Findings}

This study utilized a 3D CNN model within deep learning to investigate the applicability of non-contrast chest CT scans for identifying the presence of PE. The following are the research findings obtained during the experimental process:

\begin{enumerate}
\item Impact of HU ranges: The experimental results demonstrated a significant influence of different HU ranges on the performance of the deep learning model. 
Performance in detecting true negatives for the latter set of HU ranges was moderately satisfactory, indicating that adjusting the HU range is one of the factors affecting model performance.
\item Model advantages and limitations: The deep learning model employed in this study strikes a balance between complexity and simplicity, making it capable of achieving good model performance under limited computational resources. Simultaneously, the model is not overly complex, which helps mitigate the risk of overfitting. However, this model heavily relies on the quality and diversity of the dataset, and a limited dataset may result in reduced generalization performance.
\end{enumerate}



\subsection{Future Research Suggestions}
There are several promising directions for future research to enhance the model's performance and practical applications.
\begin{enumerate}
\item Explore different HU ranges: As evidenced by this study's experimental results, different HU ranges have a notable impact on the model's performance. Future research can further explore the effects of various HU ranges on model performance to enhance classification accuracy. This exploration could help identify HU ranges that are better suited for non-contrast chest CT scans.
\item Multimodal fusion: In addition to HU ranges, other image features like shape, texture, and structure can be considered for integration. Employing multiple features can enhance the model's ability to classify non-contrast chest CT images for PE.
\item Multimodal learning: Future research can explore the integration of different modalities of medical data, such as the combination of physiological parameters and clinical information. The incorporation of multimodal learning can provide a more comprehensive basis for diagnosis and further improve the model's accuracy.
\end{enumerate}




\end{document}